\documentclass{article}

\usepackage[preprint]{neurips_2019}

\usepackage[utf8]{inputenc} 
\usepackage[T1]{fontenc}    
\usepackage{hyperref}       
\usepackage{url}            
\usepackage{booktabs}       
\usepackage{amsfonts}       
\usepackage{nicefrac}       
\usepackage{microtype}      

\usepackage{epsfig}
\usepackage{graphicx}
\usepackage{epstopdf}
\usepackage[ruled]{algorithm2e}
\usepackage{algpseudocode}
\usepackage{graphicx}
\usepackage{epstopdf}
\usepackage{url}
\usepackage{epsfig}
\usepackage{amsmath,amssymb}
\usepackage{multirow}
\usepackage{caption}
\usepackage{fancyvrb}
\usepackage{mathptmx}
\usepackage{alltt}
\usepackage{amsmath}
\usepackage{comment}
\usepackage{amsfonts}
\usepackage{booktabs}
\usepackage{tabularx}
\usepackage{float}
\usepackage{array}
\usepackage{subcaption}
\usepackage[english]{babel}

\usepackage{natbib}
\setcitestyle{square,numbers}

\title{Generative Adversarial Networks for Mitigating Biases in Machine Learning Systems}

\author{
  Adel Abusitta\thanks{Department of Computer Science and Operations Research, E-mail: adel.abu.sitta@umontreal.ca}\\
  University of Montreal\\
  Montreal, Canada\\
  \And
  Esma A\"{i}meur\thanks{Department of Computer Science and Operations Research, E-mail: aimeur@iro.umontreal.ca}\\
  University of Montreal\\
  Montreal, Canada\\
  \And
  Omar Abdel Wahab\thanks{Department of Computer Science and Engineering, E-mail: omar.abdulwahab@uqo.ca}\\
  Université du Québec en Outaouais\\
  Gatineau, Canada
}

\begin{document}

\maketitle

\begin{abstract}
In this paper, we propose a new framework for mitigating biases in machine learning systems. The problem of the existing mitigation approaches is that they are model-oriented in the sense that they focus on tuning the training algorithms to produce fair results, while overlooking the fact that the training data can itself be the main reason for biased outcomes. Technically speaking, two essential limitations can be found in such model-based approaches: 1) the mitigation cannot be achieved without degrading the accuracy of the machine learning models, and 2) when the data used for training are largely biased, the training time automatically increases so as to find suitable learning parameters that help produce fair results. To address these shortcomings, we propose in this work a new framework that can largely mitigate the biases and discriminations in machine learning systems while at the same time enhancing the prediction accuracy of these systems. The proposed framework is based on conditional Generative Adversarial Networks (cGANs), which are used to generate new synthetic fair data with selective properties from the original data. We also propose a framework for analyzing data biases, which is important for understanding the amount and type of data that need to be synthetically sampled and labeled for each population group. Experimental results show that the proposed solution can efficiently mitigate different types of biases, while at the same time enhancing the prediction accuracy of the underlying machine learning model.
\end{abstract}

\section{Introduction}
The world is facing a historical shift toward adopting Artificial Intelligence (AI) to automate the decision-making process in many sectors, including those of health, transportation and public services. This, however, has led to growing concerns about the bias and discrimination that these systems might produce, which might negatively affect citizens especially those who belong to ethnic and racial minorities. The hazard of bias becomes even more crucial when these systems are applied to critical and sensitive domains such as health care and criminal justice. In fact, biased AI systems are mainly engendered by the data used to feed the training process of the machine learning algorithms \cite{campolo2017ai}. Training data can be incomplete, insufficiently diverse, biased, and/or consisting of non-representative samples that are not well (or poorly) defined before use \cite{campolo2017ai}, which might lead to biased results and lower accuracy \cite{campolo2017ai}. Obtaining and labeling new data to compensate and overcome these problems is one possible solution to fight against biases. However, it has been shown that such a strategy is largely difficult, costly, privacy-sensitive and dangerous, especially in some critical domains like transportation and health \cite{goodfellow2016deep} \cite{kingma2014semi}.

Many approaches have been recently proposed to fight against bias and discrimination in machine learning systems. The problem of the existing mitigation approaches \cite{louppe2017learning} \cite{madras2018learning} is that they overlook the fact that the data used to train the machine learning algorithm might be the root cause of unfair results. In particular, these approaches focus on tuning the training algorithms to decrease the chances of producing biased results. Although such a model-based strategy might end up producing fair results, the accuracy of the underlying machine learning algorithm will be largely degraded. In other words, the mitigation will be achieved on the account of the overall prediction accuracy \cite{celis2019improved}. Besides, when the training data are largely biased, the time needed to complete the training and obtain a fair model will dramatically increase, compared to the case of traditional training algorithms. The reason is that these approaches not only try to minimize the loss function (in order to teach the machine learning model), but also work on minimizing the chances of producing unfair results. Thus, a longer training time is needed to find the suitable parameters for a fair model.

To address the above-mentioned shortcomings, we propose a new framework for mitigating biases in machine learning systems, without degrading their accuracy. The proposed framework is based on conditional Generative Adversarial Networks (cGANs) \cite{mirza2014conditional}, special versions of the Generative Adversarial Networks (GANs) \cite{goodfellow2014generative}, which have shown unprecedented success in generating high-quality new synthetic data with selective properties. The proposed framework allows the designers of the machine learning systems to estimate the real distribution of the original data pertaining to the targeted population groups (population groups that are victims of biases) through formulating a minimax two-player game \cite{abusitta2018trustworthy} \cite{abusitta2018trust}. The game is played between two models, which are trained simultaneously, i.e., the Discriminator ($Dis$) and the Generator ($Gen$). $Gen$ is trained to capture the data distribution through trying to maximize the probability of $Dis$ committing a mistake. On the other hand, $Dis$ is trained to maximize the probability that a data sample came from a targeted population group rather than the $Gen$. The training of both $Dis$ and $Gen$ is repeated over many iterations until a generative model that can generate new synthetic data pertaining to the targeted population groups is obtained. The resulting generative model is then used to synthetically produce new data, which are used to augment the training set so as to compensate and overcome the bias problem. In this way, machine learning algorithms can be trained on these data in order to produce unbiased predictions.

Unlike similar works (e.g., \cite{xu2018fairgan}), the proposed model gives the designers of the machine learning systems the flexibility to decide on the amount of data that needs to be synthetically sampled and labeled, taking into account their domain knowledge. The proposed framework is also designed to be integrated into another framework for analyzing and understanding data biases. The objective is to guide the machine learning model designers on the amount and type of data that needs to be synthetically sampled and labeled. This, in turn, minimizes the chances of synthetically generating unnecessary data. Our contributions are summarized as follows. \textbf{First}, we propose a new framework for mitigating biases in machine learning systems while at the same time enhancing their overall accuracy. \textbf{Second}, we integrate the proposed mitigation framework into an analytical framework for understanding data biases. This allows us to infer the type and amount of data that needs to be synthetically sampled in order to augment the training data. \textbf{Finally}, we propose a new framework that gives the designers of the machine learning systems the flexibility to decide on the amount of data that needs to be synthetically sampled and labeled, taking into account their domain knowledge.

\section{Related Work}

The idea of using adversarial training for mitigating biases in machine learning systems has recently been addressed in several works. For example, Madras et al. \cite{madras2018learning} propose a ``fair'' representation of data \cite{lecun2015deep} that can be used by the classifier to generate fair decisions. They employ GANs to ensure that the generated representation of data is fair. Similarly, Louppe et al. \cite{louppe2017learning} propose a new approach called ``Pivot-based approach''. The framework also uses GANs not to generate new synthetic data but to create a new classifier that guarantees unbiased predictions. The method modifies the GANs design through changing the role of the generator from learning how to generate new synthetic data to a classifier that is used to produce fair results. During the training of GANs, the classifier is optimized and updated based on the prediction losses of the sensitive attributes (Ethnicity, Gender, etc.). The main disadvantage of this approach is that it does not care about the overall accuracy of the classifier during the bias mitigation process. It only cares about reducing the biased results in the classifier. In other words, the mitigation in this approach is achieved on the account of the overall accuracy. In contrast, our framework can reduce biases while at the same time enhancing the overall system's accuracy.

Xu et al. \cite{xu2018fairgan} also adopt the GANs with the aim of generating new synthetic fair data, which are then used to train the classifier on how to produce unbiased decisions. For this purpose, another discriminator was used to check if the fairness has been achieved or not. Similar approaches have been proposed in \cite{calmon2017optimized}, \cite{feldman2015certifying} \cite{kamiran2012data} and \cite{krasanakis2018adaptive}. These data-driven mitigation approaches suffer from three essential shortcomings. First, they propose to generate new data for each particular population group, thus leading to unnecessary data and unnecessary overhead. Second, these approaches require frequently verifying the machine learning model to check whether the generated data lead to a fair model or not. Third, these approaches are not complemented by any framework for analyzing and understanding data biases. This makes the designers of the machine learning systems unable to efficiently estimate and understand the amount and type of data that need to be synthetically sampled and labeled.

In contrast, our proposed mitigation approach is coupled with a framework for analyzing data biases. This is important to understand the amount of data that needs to be synthetically sampled for each particular population group. Moreover, the proposed framework gives the designers of machine learning systems the flexility to decide on the amount of data that should be synthetically sampled, taking into account both the domain knowledge and prediction accuracy with respect to the original data. As a result, the proposed model enables us to achieve fair machine learning systems while at the same time enhancing the accuracy of the prediction with minimum training overhead.

Celis et al. \cite{celis2019improved} formulate the adversarial problem as a multi-objective optimization model and try to find the fair model using a gradient descent-ascent algorithm with a modified gradient update step \cite{celis2019improved}. In fact, their approach is inspired by the work proposed by \cite{zhang2018mitigating}, while adding more robust theoretical foundations. Similarly, Agarwal et al. \cite{agarwal2018reductions} propose a minimax optimization problem, which is solved using the saddle point methods \cite{kivinen1997exponentiated} in order to derive the fair model. Other model-based mitigation approaches also are proposed in \cite{goh2016satisfying} \cite{pleiss2017fairness} \cite{woodworth2017learning} \cite{hardt2016equality}. These approaches propose algorithms to find suitable thresholds for trained classifiers so as to ensure equalized and fair odds. In particular, they try to fix the decision boundary in such a way to ensure that the final classifier is fair.

Most of the above-mentioned model-based mitigation approaches do not consider the training data as a potential reason for biased results. Instead, they focus only on modifying the training algorithms to produce fair results. Two main disadvantages can be distinguished in such an approach. First, the mitigation is achieved on the account of the accuracy. Second, the time needed to obtain the fair model is higher than that in traditional training algorithms, especially when the data used for training are largely biased \cite{abusitta2019deep} \cite{abusitta2018svm}. This is because these models are not only trained to minimize the loss function, but also to minimize the chances of producing unfair results.

\section{The Proposed Framework for Mitigating Machine Learning Biases}
In this section, we provide the details of the our framework proposed for mitigating biases in machine learning systems. We first give some explanations on Generative Adversarial Networks and conditional Generative Adversarial Networks and then present the proposed mitigation model in detail, followed by our framework for analyzing data biases.

\subsection{Generative Adversarial Nets and the Conditional Version}
Generative adversarial networks (or GANs) is a new generative model that has been proposed by \cite{goodfellow2014generative}. A generative model can be seen as a way of learning any kind of data distribution using unsupervised learning techniques \cite{bengio2007greedy} \cite{hinton2006fast}. Although several generative models have been proposed in the literature such as Deep Belief Network (DBN) \cite{hinton2006fast} and Variational Autoencoder (VAE) \cite{doersch2016tutorial},  GANs have received more attention thanks to their unprecedented ability to generate new synthetic high-quality data compared to the traditional generative models. In fact, GANs consist of two models: a discriminative ($Des$) and a generative ($Gen$) models. $Gen$ is trained to capture the data distribution through trying to maximize the probability of $Dis$ committing a mistake. On the other hand, $Dis$ is trained to maximize the probability that a data sample came from a targeted population group rather than the $Gen$. The training of both the discriminative and generative models is repeated over many iterations until the discriminative model becomes unable to distinguish whether the underlying data is a sample from the data or generated from the generater. This framework is also known as a minimax two-player game \cite{o1987nonmetric} \cite{halabi2019toward} \cite{halabi2018cooperative} and is described formally as follows:

\begin{equation}\small
min_{Dis}\hspace{0.2cm}max_{Gen} V(Dis,Gen)=\mathbf{E}_{x\thicksim p_{data}(x)}log[Dis(x)] + \mathbf{E}_{z\thicksim p_{z}(z)}log[1-Dis(Gen(z))]
\end{equation}

Conditional Generative Adversarial  Networks (or cGANs) \cite{mirza2014conditional} are a special case of GANs which have shown great success in generating high-quality new synthetic data with selective properties. Although Goodfellow et. al \cite{goodfellow2014generative} have already indicated in their original work the possibility of training cGANs, their work did not provide theoretical and experimental results to support this claim. cGANs can be achieved through adding a condition $c$ as an input in both $Gen$ and $Dis$. The formal description of cGANs is described as follows:

\begin{equation}\small
min_{Dis}\hspace{0.2cm}max_{Gen} V(Dis,Gen)=\mathbf{E}_{x\thicksim p_{data}(x)}log[Dis(x|c)] + \mathbf{E}_{z\thicksim p_{z}(z)}log[1-Dis(Gen(z|c))]
\end{equation}

\subsection{The Proposed Model}
The proposed mitigation model is based on cGANs. In particular, we train  $Gen$ to synthetically produce new synthetic data based on the Targeted Population Groups ($TPG$). $TPGs$ represent those population groups against whom the machine learning models produce biased results. The new data generated using the proposed framework are then used to augment the training data (incomplete and biased data). The new data (original data and generated data) will then be used to train the machine learning algorithms.  Figure 1 depicts the architecture of our proposed model.

In the next section, we present a new framework used for analyzing data biases and exploring the $TPGs$. This framework is designed to be integrated into the proposed mitigation approach in order to allow the designers of the machine learning systems to understand the amount and type of data that should be synthetically sampled for each population group. To this end, the objective function of a two-player minimax game is defined as follows:

\begin{equation}\small
min_{Dis}\hspace{0.2cm}max_{Gen} V(Dis,Gen)=\mathbf{E}_{x\thicksim p_{data}(x)}log[Dis(x|TPG)] +\mathbf{E}_{z\thicksim p_{z}(z)}log[1-Dis(Gen(z|TPG))]
\end{equation}

\begin{figure}[htbp]\small
\centering
\scalebox{0.40}{\includegraphics{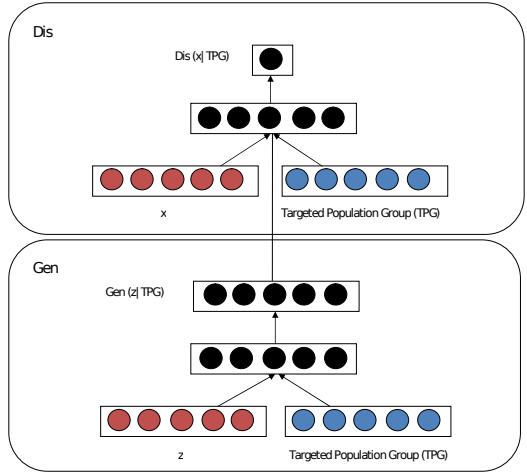}}
\caption{The architecture of our proposed model}
\label{fig3}
\end{figure}

Since the standard training of GANs cannot easily converge (i.e., non-convergence problem) \cite{goodfellow2016nips} and to avoid mode collapse \cite{goodfellow2016nips}, we adopt a Primal-Dual Sub-gradient method to solve this problem. This method is proposed by \cite{chen2018training} and can be seen as a Lagrangian perspective of GANs \cite{chen2018training}. To this end, we construct a convex optimization problem as follows:
\begin{equation}\small\small
\begin{aligned}
&maximize\hspace{0.1cm} \sum_{i=1}^{n}p_{data}(x_{i}|TPG)log(Dis(x_{i}|TPG))\\
&Subject\hspace{0.1cm}to:\hspace{0.1cm}(1-log(Dis(x_{i}|TPG))\geq log(1/2), i= 1, ..., n\\
&\hspace{1.8cm}\textbf{\emph{Dis}}\in\mathcal{S},
\end{aligned}
\end{equation}
where $\mathcal{S}$ is some convex set and the variables are \textbf{\emph{Dis}}=($Dis$($x_{1}$|$TPG$),...$Dis$($x_{n}$|$TPG$)). Let $p_{gen|TPG}$ = ($p_{gen}$($x_{1}$|$TPG$), ... , $p_{gen}$($x_{n}$|$TPG$)), where $p_{gen}$($x_{i}$|$TPG$) is the Lagrangian dual associated with the $i$-th constraint. Therefore, the Lagrangian function becomes as follows:
\begin{equation}\small
L(\textbf{\emph{Dis}},p_{gen})= \sum_{i=1}^{n}p_{data}(x_{i}|TPG)log(Dis(x_{i}|TPG))+ \sum_{i=1}^{n}p_{gen}(x_{i}|TPG)log(2(1-Dis(x_{i}|TPG))
\end{equation}
The proposed training algorithm (Algorithm 1), which is inspired by \cite{chen2018training}, is based on (5). In Algorithm 1, the targeted population group ($TPG$) is taken as an input and the goal is to train $Gen$ to produce data that cope with the $TPG$. In the proposed algorithm, the process of updating of $Dis$ is similar to the standard $cGAN$ training; however, the process of updating $Gen$ is different. For the $Gen$, when the data distribution and generated distribution have disjoint supports \cite{gulrajani2017improved} \cite{chen2018training}, the $Gen$ may not be updated using standard $cGAN$ training (7) (8) (9). This is useful to prevent the main source of mode collapse \cite{chen2018training}. Note that after a certain fixed period of time denoted by $\varepsilon$, the whole steps are repeated in order to enable both the $Gen$ and $Dis$ to learn how to produce new high-quality synthetic data, based on the targeted population group.
\begin{algorithm}[t]\small
\SetKwInput{Initialization}{Initialisation}
\LinesNumbered
\SetKwInOut{Input}{Input}\SetKwInOut{Output}{Output}
Input: Targeted Population Group (TPG)

\Repeat{$\varepsilon$ elapses}
{

Sample n1 data samples $x_{i}$, $i$= 1, ..., n1  (minibatch sampling) \\
Sample n2 noise samples $z_{i}$, $i$= 1, ..., n2 (minibatch sampling) \\
\For{K steps}
{
Update the $Dis$ through ascending the stochastic gradient:
\begin{equation}
\nabla_{\theta_{data}}[\frac{1}{n1}\sum_{1}^{n1}log(Dis(x_{i}|TPG)) + \frac{1}{n2}\sum_{1}^{n2} log(1-Dis(Gen(z_{i}|TPG)))]
\end{equation}

}
Update the $Gen$ distribution as follows:
\begin{equation}
\tilde{p}_{gen}(x_{i}|TPG)= p_{gen}(x_{i}|TPG) - \beta log(2(1-Dis(x_{i}|TPG))), i= 1,...n1
\end{equation}
where $\beta$ represents some step size and

\begin{equation}
p_{gen}(x_{i}|TPG) = \frac{1}{n2}\sum_{j=1}^{n2} k_{\sigma}(Gen(z_{j}|TPG)-x_{i}).
\end{equation}
Update the Gen through descending the stochastic gradient:
\begin{equation}
\nabla_{\theta_{gen}}[\frac{1}{n2}\frac{1}{n2}log (1-Dis(Gen(z_{j}|TPG))) + \frac{1}{n1}\sum_{1}^{n1}(\tilde{p}_{gen}(x_{i}|TPG) - p_{gen}(x_{i}|TPG))^{2}]
\end{equation}

}
\caption{Algorithm for training a generator}
\end{algorithm}
\subsection{A Framework for Analyzing Data Biases}
In the previous section, we proposed a new algorithm (Algorithm 1) for learning how to train the generator on how to create new synthetic data based on a given targeted population group. The algorithm takes as an input a targeted population group in order to learn how to produce new data with respect to that particular group. In this section, we present a new framework that can be used to explore the set of targeted population groups to be used as inputs for Algorithm 1. Note that this framework is inspired by the analysis presented in \cite{biaslink} for detecting biases in machine learning models, while adapting it to our case where we are interested in detecting biases in the data itself rather than in the machine learning model.

The following steps are used for the analysis of data biases. First, select a set of population groups to study if the classifier produces biased results against any of them. Second, train the classifier on the training data. Third, test the classifier by producing results and visualizing the prediction accuracy with respect to each population group. The visualization can be achieved either by showing the probability distribution or by displaying the accuracy obtained for each population group. Finally, analyze these results to see which population group(s) is/are victim(s) of biases.

We use the following example to illustrate how does the above-described framework practically work. Consider the adult UCI dataset \cite{yen2006under}, which is used to predict the salary of a person (below 50K\$ or above 50k\$). The dataset contains two Sensitive Attributes (\emph{SA}), i.e., Ethnicity and Gender. This leads us to the four following population groups:  African American, Caucasian, Female and Male. Although we could have combinations of these population groups (e.g., African American females), we restrict, for the sake of simplicity and without loss of generality, our example to only the above mentioned four population groups.

To determine if the training data are biased or not, we need to test whether a machine learning classifier, that is trained on these data, produces biased results or not. To this end, we trained a neural network classifier on this dataset and analyzed the prediction accuracy, taking into account above mentioned population groups. The results of our testing are given in Figure 2.

Figure 2a shows the distributions of the predicted $P$(income $>$ 50K\$ ) given the \emph{SA} $S_{Ethnicity}$ = $\{$African American, Caucasian$\}$. The Figure shows that for the ethnicity attribute, the prediction distribution of an ``African American'' has a large value at the low interval of $[0.1 -0.2]$ compared to a ``Caucasian''. These results suggest that when a person is an ``African American'', the probability that the classifier will predict his/her income below 50K\$ is much higher compared to a ``Caucasian''. Similarly, Figure 2b shows the distributions of the predicted $P$(income $>$ 50K\$) given the \emph{SA} $S_{Gender}$ = $\{$female,male$\}$. The Figure shows that for the gender attribute, the prediction distribution of  a ``female'' has a large value at the low interval of [0.1 -0.2] compared to a ``male''. These results suggest that when a person is ``female'', the probability that the classifier will predict her income below 50 K\$ is much higher compared to a ``male''.
\begin{figure*}[t]\small
	\begin{subfigure}{0.48\textwidth}
		\centering
		\captionsetup{width=.8\linewidth}
		\includegraphics[width=0.9\linewidth]{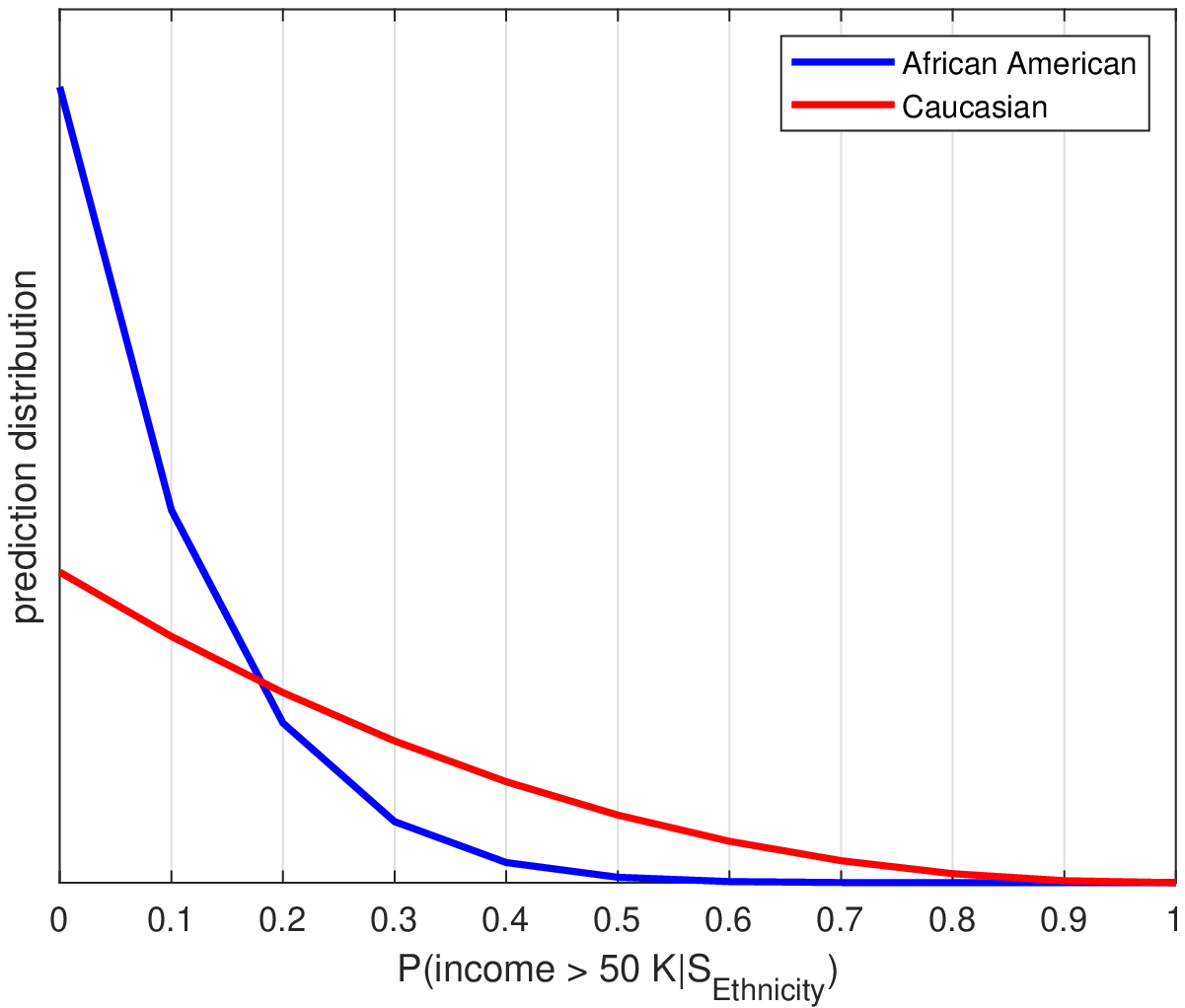}
		\caption{}
		\label{figure:2a}
	\end{subfigure}%
	\begin{subfigure}{0.48\textwidth}
		\centering
		\captionsetup{width=.8\linewidth}
		\includegraphics[width=.9\linewidth]{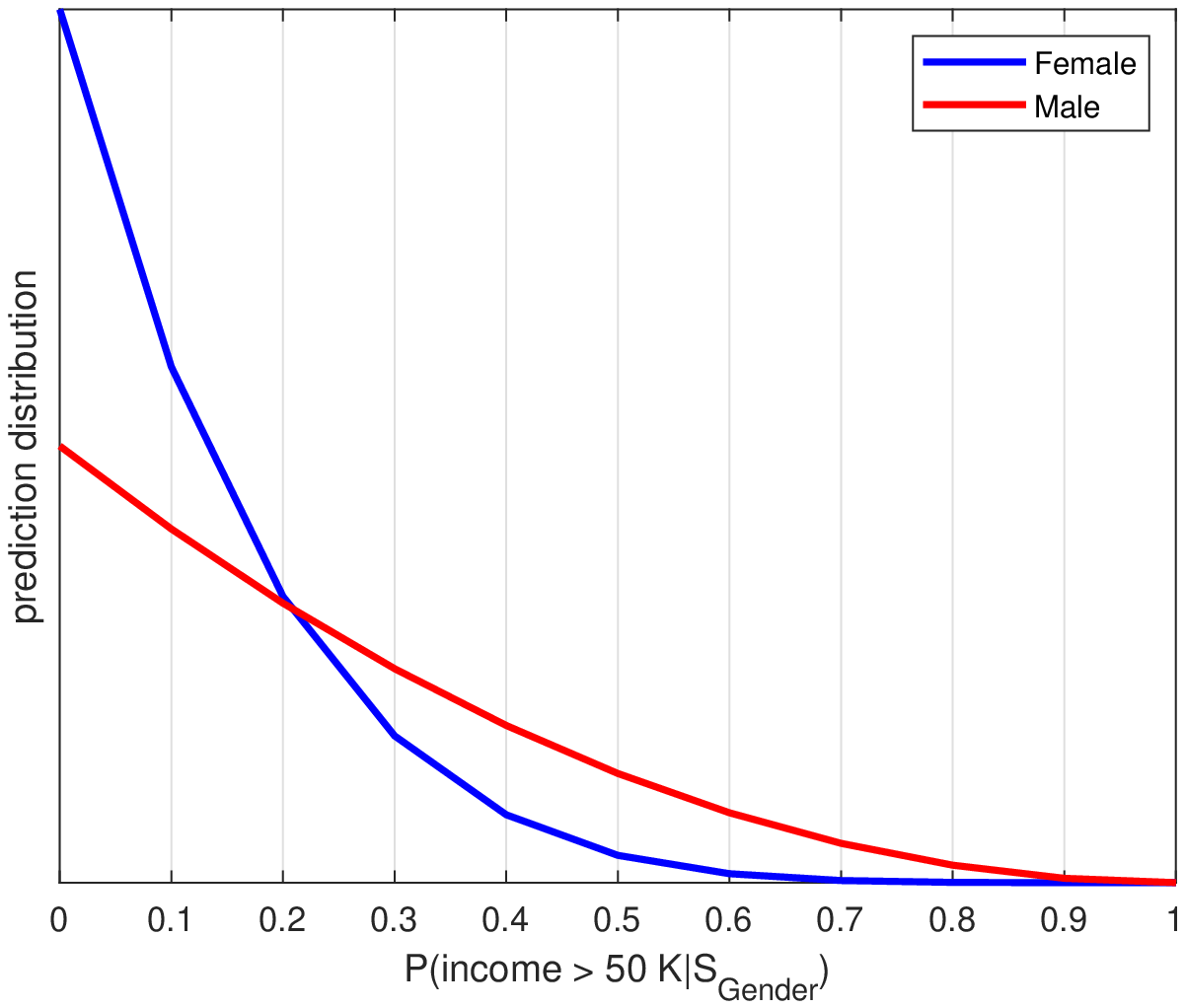}
		\caption{}
		\label{figure:2b}
	\end{subfigure}
	\caption{(Left) Prediction distribution of the original training data with respect to the ethnicity attribute. (Right) Prediction distribution of the original training data with respect to the gender attribute.}	\vspace{-4 mm}
	\label{figure:2}
\end{figure*}
The results shown in Figure 2 give us a clear indication that the data used for training is incomplete (i.e., the number of Caucasians and males in the dataset is greater than that of African Americans and females). Therefore, we conclude that the targeted population groups that should be used as inputs to Algorithm 1 based on to the above results are:  $S_{Ethnicity}$ = $\{$African American$\}$ and $S_{Gender}$ = $\{$female$\}$. Simply put, the generator will be trained to generate new African Americans and females.

\section{Experimental Evaluation}
This section first describes the setup used to evaluate the proposed framework. Then, the performance of the proposed bias mitigation framework is examined.

\subsection{Experimental Setup}
We implemented the proposed framework using Multilayer Perceptrons (MLPs) with 3 hidden layers. We used the ReLU activation function for both the generators and discriminators. The two following datasets were tested: the adult UCI dataset \cite{yen2006under} and the Adience dataset \cite{adience2019}, which widely used for age and gender prediction. Since the adult UCI dataset contains categorial data, we placed in parallel a dense-layer per categorical variable, followed by Gumbel-Softmax activation and a concatenation to get the final output \cite{camino2018generating} \cite{jang2016categorical} \cite{maddison2016concrete}. Prediction performance on the validation dataset is adopted for finding the best hyper-parameter configuration. The results are reported based on a 95\% confidence interval.

\subsubsection{Results on the adult UCI dataset}
Figure 3 shows the results obtained when applying the proposed framework on the adult UCI dataset. In particular, Figure 3a shows the progress achieved in the prediction distribution compared to Figure 2a. This progress was achieved when we augmented the original data (female) by 85\% new data obtained synthetically from the generator.  Figures 3b also shows the progress achieved in the prediction distribution, compared to Figure 2b,  when we augmented the original data (African American) by 85\% new data obtained synthetically from the generator. Note that the proposed framework is flexible in the sense that it enables machine learning designers to control the amount of data (e.g, 85\%) that needs to be synthetically added for each population group. This allows the designers to consider the ``Domain knowledge'' during the data augmentation process.

\begin{figure*}[t]\small
	\begin{subfigure}{0.48\textwidth}
		\centering
		\captionsetup{width=.8\linewidth}
		\includegraphics[width=0.9\linewidth]{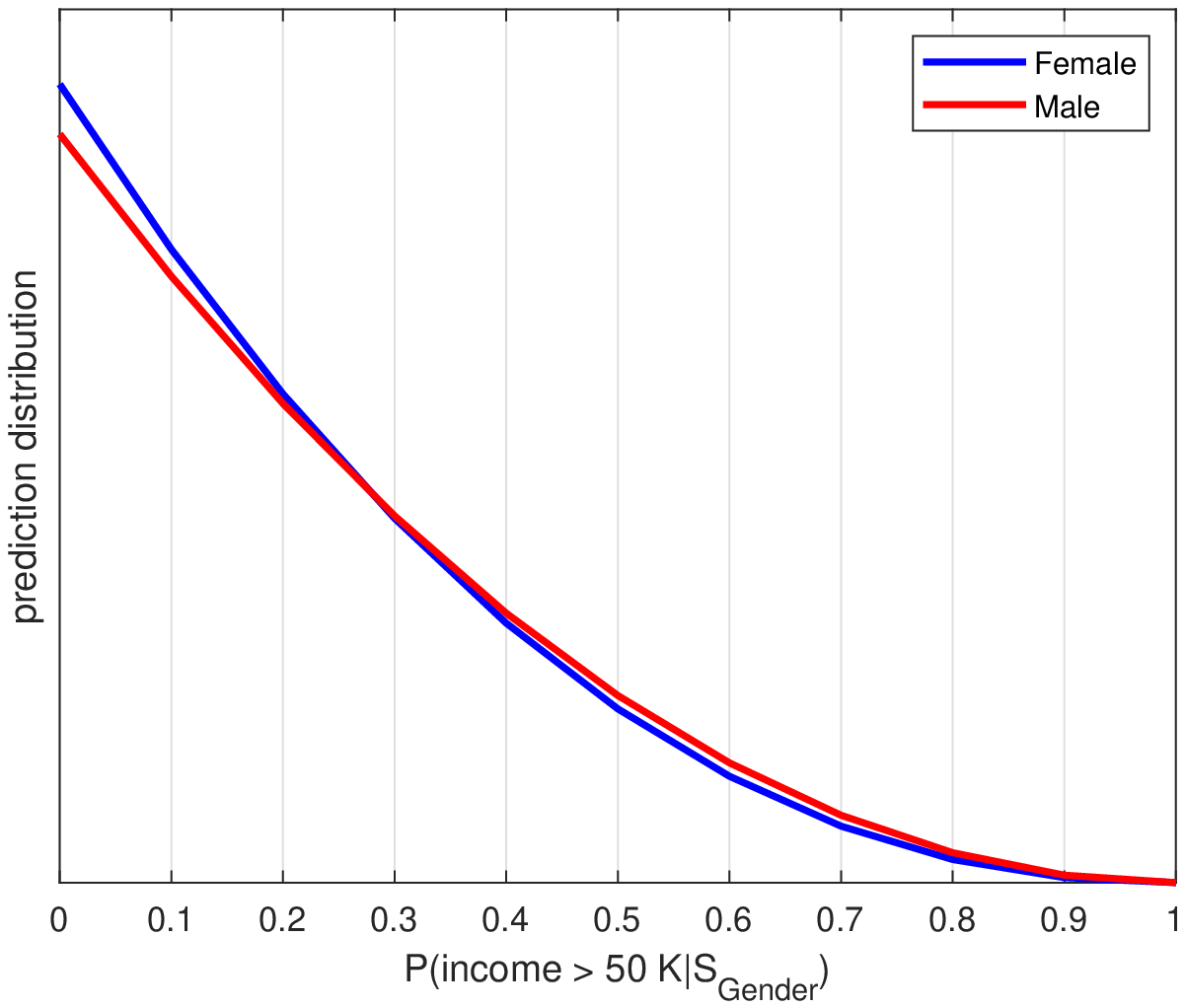}
		\caption{}
		\label{figure:2a}
	\end{subfigure}%
	\begin{subfigure}{0.48\textwidth}
		\centering
		\captionsetup{width=.8\linewidth}
		\includegraphics[width=.9\linewidth]{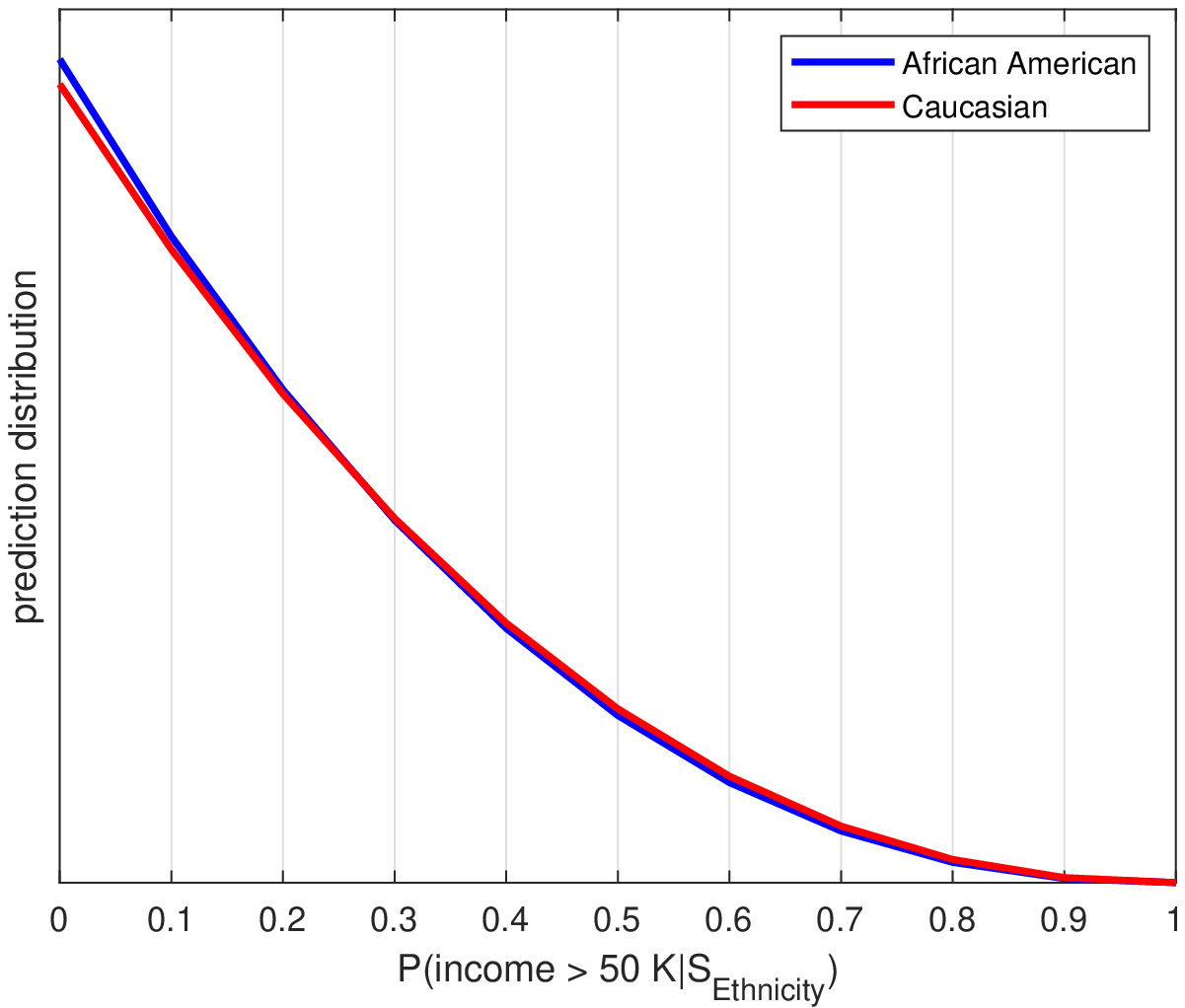}
		\caption{}
		\label{figure:2b}
	\end{subfigure}
	\caption{(Left) Prediction distribution when 85\% of new synthetic data (female) were added to the original dataset. (Right) Prediction distribution when 85\% of new synthetic data (African American) were added to the original dataset.}	\vspace{-4 mm}
	\label{figure:2}
\end{figure*}

Table 1 shows a comparison between the proposed approach and a recent work proposed in \cite{louppe2017learning}. This work is called as a ``Pivot-based mitigation approach'' and it uses GANs not to generate new synthetic data (like we do) but to create a new classifier that guarantees fairness in predictions. The method makes a modification on the GANs through changing the role of the generator from learning how to generate new synthetic data to a classifier that is used to produce fair results. During the training process of GANs, the classifier is optimized and updated based on the prediction losses of the sensitive attributes (e.g., Ethnicity, Gender, etc.).

Table 1 shows the overall accuracy obtained by the proposed model when training the MLP on the new training data (original data + generated data) with different numbers of Hidden Units (HUs). These results are better than the results obtained using the `Pivot-based mitigation approach''. Our model also yields a better accuracy compared to the baseline. The baseline means that the classifier was trained on the original data without adding new synthetic data.  This can be justified by the fact that the data used for training was incomplete and led to biased results, in the sense of having a lower measure of accuracy \cite{campolo2017ai}. The proposed framework overcame this problem through augmenting the training data to mitigate biases and enhancing the prediction accuracy.
\begin{table}[ht]\small\small\small
\caption{Comparison of the prediction accuracy of different approaches (Adult UCI dataset)}
\centering
 \begin{tabular}{c c c c c}
 \hline
 \textbf{ }& \textbf{Acc. (300 HUs)}& \textbf{Acc. (500 HUs)}& \textbf{Acc. (700 HUs)}& \textbf{Acc. (900 HUs)}\\
\hline
 \textbf{The Proposed Approach} & \textbf{84.9$\pm$ 1.14} & \textbf{85.1$\pm$ 1.09} & \textbf{85.3$\pm$ 1.92} & \textbf{85.5$\pm$ 1.15}\\
 \textbf{Pivot-based Approach} & 76.1 $\pm$ 1.11 & 76.4 $\pm$ 1.84 & 77.1$\pm$ 1.23 & 77.3$\pm$ 1.78\\
 \textbf{Baseline}  & 82.0$\pm$ 1.16 & 82.3$\pm$ 1.06 & 82.6 $\pm$ 1.90& 82.9 $\pm$ 0.88
\end{tabular}
\end{table}

\subsubsection{Results on the Adience dataset}

Table 2 studies the accuracy of the MLP classifier with respect to a given population group. The results suggest the existence of bias against the women of color. Table 3 shows the progress achieved in the prediction accuracy compared to Table 2 when the training data was augmented with more data on women of color, which were synthetically obtained from the generator (the proposed framework).

\begin{table}[ht]\small\small\small
\caption{Classification performance with respect to a population group}
\centering
 \begin{tabular}{c c c c c}
 \hline
 \textbf{ }& \textbf{Acc. (300 HUs)}& \textbf{Acc. (500 HUs)}& \textbf{Acc. (700 HUs)}& \textbf{Acc. (900 HUs)}\\
\hline
 \textbf{Men of color} & 86.5 $\pm$ 0.34 & 87.68 $\pm$ 0.33 & 87.15 $\pm$ 0.22 & 87.5 $\pm$ 0.26\\
 \textbf{Women of color} &\textbf{67.8$\pm$ 0.14} & \textbf{67.9$\pm$ 0.29} & \textbf{68.3$\pm$ 0.36} & \textbf{68.4 $\pm$ 2.40}\\
 \textbf{Caucasian men}  & 98.6 $\pm$ 0.24 & 98.7 $\pm$ 0.35 & 98.8 $\pm$ 0.19 & 98.0 $\pm$ 0.21\\
 \textbf{Caucasian women} & 90.4 $\pm$ 0.45 & 91.3 $\pm$ 0.38 & 91.8 $\pm$ 2.02 & 91.7 $\pm$ 0.44
\end{tabular}
\end{table}
\begin{table}[ht]\small\small
\caption{Classification performance after the augmentation with the data on women of color (300\%)}
\centering
 \begin{tabular}{c c c c c}
 \hline
 \textbf{ }& \textbf{Acc. (300 HUs)}& \textbf{Acc. (500 HUs)}& \textbf{Acc. (700 HUs)}& \textbf{Acc. (900 HUs)}\\
\hline
 \textbf{Men of color} & 87.9 $\pm$ 0.38 & 88.1 $\pm$ 0.45 & 88.1 $\pm$ 0.84 & 88.3 $\pm$ 0.66\\
 \textbf{Women of color} &\textbf{88.1 $\pm$ 0.27} & \textbf{88.2 $\pm$ 0.30} & \textbf{88.5 $\pm$ 0.34} & \textbf{88.6 $\pm$ 0.22}\\
 \textbf{Caucasian men}  & 99.2$\pm$ 0.31  & 99.3 $\pm$ 0.42 & 99.5 $\pm$ 0.28 & 99.7 $\pm$ 0.37\\
 \textbf{Caucasian women} & 91.9 $\pm$ 0.66 & 92.0 $\pm$ 0.41 & 92.3 $\pm$ 2.07 & 92.5 $\pm$ 0.23
\end{tabular}
\end{table}

Table 4 shows the overall prediction accuracy of the MLP classifier trained on the new training data. These results outperform both the pivot-based classifier and the baseline.
\begin{table}[!t]\small\small\small
\caption{Comparison of overall prediction accuracy (Adience dataset)}
\centering
 \begin{tabular}{c c c c c}
 \hline
 \textbf{ }& \textbf{Acc. (300 HUs)}& \textbf{Acc. (500 HUs)}& \textbf{Acc. (700 HUs)}& \textbf{Acc. (900 HUs)}\\
\hline
 \textbf{The Proposed Approach} & \textbf{91.77$\pm$ 0.29} & \textbf{91.9$\pm$ 0.36} & \textbf{92.10$\pm$ 0.41} & \textbf{92.27$\pm$ 0.25}\\
 \textbf{Pivot-based Approach} & 81.71$\pm$ 0.29 & 80.43$\pm$ 0.38 & 81.01$\pm$ 0.31 & 81.37$\pm$ 0.29\\
 \textbf{Baseline}  & 85.82$\pm$ 0.33 & 86.39$\pm$ 0.24 & 86.51$\pm$ 0.31 & 86.40$\pm$ 0.37
\end{tabular}
\end{table}
\section{Limitation}
Although the proposed framework has the advantage of mitigating bias in machine learning systems against targeted groups, we cannot claim that our solution fully solves the problem. In fact, bias is a broad and undefined problem, which does not always target members of minority groups (e.g., female). For example, Google conducted a recent study to determine whether the company is underpaying women or not. Surprisingly, they found that men were less paid than women even for the same job position \cite{workgoogle2019}. Therefore, we argue that more efforts need to be done to generalize the proposed framework for unpredictable bias cases.



\section{Conclusion and Future Work}
\label{sec:Conclusion}
This paper presents a new framework for the mitigation of biases in machine learning systems. The proposed framework is based on conditional generative adversarial networks, which allows us to generate new high-quality synthetic data related to the targeted population groups. The proposed framework is integrated into another analytical framework used for understanding of data biases. This allows us to understand the type and amount of data that should be synthetically sampled to augment the training data and overcome the bias problem. The training process then takes place on the new data (original data + generated data). Our model also enables the mitigation to be applied while taking into consideration the knowledge domain. Experimental results show that the proposed framework mitigates the biases against targeted population groups while at the same time enhancing the prediction accuracy of the machine learning classifiers.

As future work, we plan to design an automated mitigation process. In particular, after defining the bias, the system should automatically generate new data and perform unbiased training. The challenge here is to make the system automatically determine the exact amount of data that should be sampled, taking into account the knowledge domain.

\section*{Acknowledgment}
The financial support of the Natural Sciences and Engineering Research Council of Canada is gratefully acknowledged. We also would like to acknowledge Dr. Gilles Brassard (University of Montreal), Dr. Kimiz Dalkir (McGill University), Younes Driouiche (Mila), Alexis Tremblay, Amine Belabed and Rim Ben Salem for helpful discussions.



\bibliographystyle{abbrvnat}
\bibliography{references}

\end{document}